# TALplanner in the Third International Planning Competition: Extensions and Control Rules

**Jonas Kvarnström**                                                                 JONKV@IDA.LIU.SE
**Martin Magnusson**                                                                 MARMA@IDA.LIU.SE
*Department of Computer and Information Science*
*Linköping University*
*SE-581 83 Linköping, Sweden*

## Abstract

TALplanner is a forward-chaining planner that relies on domain knowledge in the shape of temporal logic formulas in order to prune irrelevant parts of the search space. TALplanner recently participated in the third International Planning Competition, which had a clear emphasis on increasing the complexity of the problem domains being used as benchmark tests and the expressivity required to represent these domains in a planning system. Like many other planners, TALplanner had support for some but not all aspects of this increase in expressivity, and a number of changes to the planner were required. After a short introduction to TALplanner, this article describes some of the changes that were made before and during the competition. We also describe the process of introducing suitable domain knowledge for several of the competition domains.

## 1. Introduction

Like most planners, TALplanner (Kvarnström & Doherty, 2000; Doherty & Kvarnström, 1999; Kvarnström, Doherty, & Haslum, 2000; Doherty & Kvarnström, 2001; Kvarnström & Doherty, 2003; Kvarnström, 2002) allows the user to specify a goal in the shape of a set of atemporal logic formulas that must be satisfied in the final state that results from executing a plan. Unlike most planners TALplanner also allows the specification of a set of *temporal* logic formulas that must be satisfied by the entire *state sequence* generated by a plan.

Obviously, these formulas can be used to specify temporally extended goals, such as safety and maintenance goals that must be upheld throughout the execution of a plan. However, it is also possible to specify constraints related to traditional measures of plan quality, such as constraints that forbid certain "stupid" actions from taking place, as in the Blockhead blocks world planner by Kibler and Morris (1981) or TLPLAN by Bacchus and Kabanza (2000), which initially inspired the development of TALplanner. For example, in a logistics domain one may specify a temporally extended goal stating that once a package is at its destination, it is never picked up again, and a goal stating that trucks driving between two locations always use the shortest path. Such constraints can then be processed by TALplanner in order to automatically extract control knowledge that can be used during a forward-chaining search process, as opposed to being used as a filter after a candidate plan has been generated. Given sufficiently strong constraints, the planner can efficiently prune most of the search tree, making it easier to find a plan among the remaining nodes. Often (as in this article) the search control aspect is in fact the primary reason for introducing a temporally extended goal, in which case the goal is usually referred to as a control rule.

Although forward-chaining planners may sometimes suffer from a lack of goal-directedness when compared to other types of planners, the use of explicitly represented domain-dependent knowledge is one way of compensating for this deficiency. More significantly, a forward chaining planner always has a complete description of the past and current states, which facilitates the use of complex operator types with complex preconditions and conditional effects. This expressivity was useful when TALplanner participated in the third International Planning Competition (IPC-2002[1]), which had a clear emphasis on increasing the complexity of the problem domains used as benchmark tests and the expressivity required to represent these domains in a planning system. In fact, TALplanner already had support for several new features that had not been present in IPC-2000[2], such as the use of numeric state variables and temporally extended actions with variable duration.

Nevertheless, several extensions and changes had to be implemented before and during the competition in order to accommodate the semantics of PDDL2.1, the new version of PDDL (Planning Domain Definition Language, Fox & Long, 2003) which was used to specify problem domains and problem instances. These extensions and changes are the first topic of this article, and after an introduction to TALplanner (Sections 2 and 3), the extensions will be discussed in Section 4. The second topic is that of describing the domain-dependent control rules that were used for the six benchmark problem domains in the hand-tailored track of the competition, and more importantly, the process of generating those rules and the reasoning behind them (Section 5). We will also describe some new changes that have been made to TALplanner after the competition (Section 6). Finally, we will conclude with a discussion of the positive and negative sides of using search control knowledge in TALplanner together with some pointers towards possible future research topics.

Please see Long and Fox (2003) for further information about the basic setup of the competition, detailed descriptions of the planning domains being used, and timing and plan quality results.

## 2. Representation: Using TAL in TALplanner

The semantics of TALplanner is based on an extended version of TAL-C (Karlsson & Gustafsson, 1999; Doherty, Gustafsson, Karlsson, & Kvarnström, 1998), a member of the TAL (Temporal Action Logics) family of narrative-based non-monotonic linear discrete metric time logics for reasoning about action and change. TAL-C has been developed for modeling domains that may include the use of incomplete information, delayed effects of actions, finite or infinite chains of indirect effects, interacting concurrent actions, and independent processes not directly triggered by action invocations. Consequently, it was seen as an ideal choice not only for the initial version of TALplanner but also for most extensions that could conceivably be implemented in the foreseeable future.

A TAL narrative consists of a set of labeled statements in a high-level macro language $\mathcal{L}(\text{ND})$, where the basic language has a number of statement classes for observations of fluent values (labeled obs), action descriptions (acs), action occurrences (occ), domain constraints (dom), and dependency constraints modeling causal relations and indirect effects (dep). The formal semantics of $\mathcal{L}(\text{ND})$ is defined by a translation into an order-sorted first-order base language $\mathcal{L}(\text{FL})$ and by a circumscription policy providing a solution to the frame and ramification problems (Doherty, 1994; Gustafsson & Doherty, 1996; Doherty et al., 1998).

---

1. http://www.dur.ac.uk/d.p.long/competition.html
2. http://www.cs.toronto.edu/aips2000/

The $\mathcal{L}(ND)$ language is designed to be easily extended for different tasks, such as planning. An extension may take the shape of a new specialized macro or a new type of statement. As illustrated in Figure 1, a TALplanner *goal narrative* uses a version of $\mathcal{L}(ND)$ called $\mathcal{L}(ND)^*$, which contains some of the standard classes of $\mathcal{L}(ND)$ statements together with several new types of planning-related statements. These extensions are accompanied by extensions to the translation function, so that the new variation of TAL can still share the same base language $\mathcal{L}(FL)$.

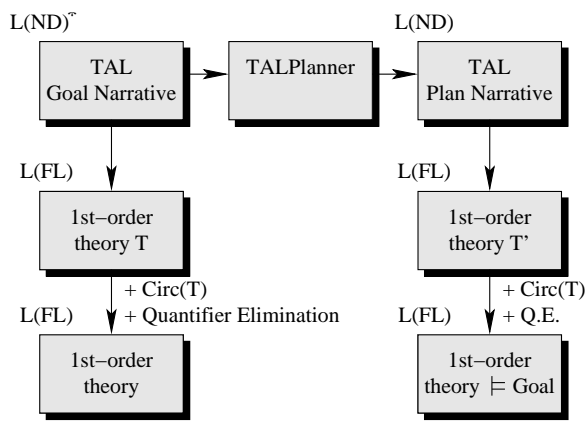

Figure 1: TAL/TALplanner relation

However, TALplanner does not use this translation directly during the planning process. Instead, it makes direct use of the higher level $\mathcal{L}(ND)^*$ goal narrative in a forward-chaining search process and generates a *plan narrative* where a set of timed action occurrences (corresponding to a plan) has been added, and where the goal is entailed in the final state.

In this section, we will attempt to provide an intuitive understanding of TAL and how it is used in domain specifications using concrete examples from the standard logistics planning domain, where a set of objects (packages) can be transported by truck between locations in the same city and by airplane between airports in different cities. The next section contains further information about the search process and the use of control rules. See Doherty et al. (1998) for a more detailed description of TAL, and see Kvarnström and Doherty (2000) for more information about TALplanner.

**Notation.** All formulas and $\mathcal{L}(ND)$ statements below will be shown using the input syntax for TALplanner, with the exception of some connectives and quantifiers that may be written using the ordinary logical symbols for increased clarity. All free variables are implicitly universally quantified.

### 2.1 Types, Objects and State Variables

Although some planners are restricted to declaring an unstructured set of objects and representing types as unary predicates, TAL is order-sorted and allows the user to specify a hierarchy of object types (sorts). The logistics domain can be modeled using the standard sort boolean = {true, false} together with the seven user-specified types: loc (location) has the subtypes airport and city, while thing has the subtypes obj and vehicle, the latter of which has the subtypes truck and plane.

TALplanner also allows the use of numeric types. In order to keep the semantics of these types clear, only integers and fixed point numbers (that is, numbers with a fixed number of decimals) are allowed, and lower and upper bounds must be declared for each numeric type. All of the standard arithmetic operators are available for the numeric types and are given an interpretation through semantic attachment.

State variables are represented using TAL fluents, which are not restricted to being predicates but can take values from an arbitrary user-specified sort. For the logistics domain, one could use two boolean fluents, at(thing, loc) and in(obj, vehicle), together with a city-valued fluent city_of(loc) denoting the city containing the location loc.

## 2.2 The Initial State

Given the fluents that were defined above, the initial state of a logistics problem instance can be specified using $\mathcal{L}(\text{ND})$ observation statements:

#obs [0] city_of(pos1) $\hat{=}$ city1 $\wedge$ city_of(pos2) $\hat{=}$ city2 $\wedge$ ...
#obs [0] at(obj11, pos1) $\wedge$ at(truck1, pos1) $\wedge$ ...

These observations consist of TAL-C *fixed fluent formulas*, formulas of the form $[\tau]\,\phi$ denoting the fact that the *fluent formula* $\phi$ holds at time $\tau$. A fluent formula is a boolean combination of elementary fluent formulas of the form $f \hat{=} v$ (f==v in input notation), denoting the fact that the fluent $f$ takes on the value $v$. For boolean fluents, as in the second observation, the shorthand notation $f$ or $\neg f$ (!f in input notation) is allowed. The notation is also extended for open, closed, and semi-open temporal intervals. In addition to these formulas, the function $value(\tau, f)$ denotes the value of $f$ at time $\tau$.

## 2.3 The Goal: Goal Statements and Goal Expressions

A statement class for goals (labeled goal) has been added to $\mathcal{L}(\text{ND})^*$. A goal statement consists of a fluent formula that must hold in any goal state:

#goal at(obj11, airport1) $\wedge$ at(obj23, pos1) $\wedge$ ...

The ability to test whether a formula is entailed by the (state-based) goal is very useful in temporally extended goals and domain-dependent control rules. Therefore, a new macro is added: The *goal expression* goal($\phi$) holds iff the goal of this problem instance (the conjunction of all goal statements) entails the fluent formula $\phi$. Stated differently, goal($\phi$) is true if $\phi$ *must* be true in *every* goal state. The translation into $\mathcal{L}(\text{FL})$ is somewhat complex; see Kvarnström and Doherty (2000) for further information.

Note that a valid plan must end in a goal state. It is not sufficient to visit a goal state temporarily, which could be the case when an operator has effects at multiple timepoints – first satisfying the goal and then destroying it – or when concurrent plans are being created. (If such plans were desired for some reason, it would of course be easy to modify the definition and the planner accordingly.)

## 2.4 Operator Definitions

Since TAL-C is a logic for reasoning about action and change, it has a notion of actions that can be used for modeling planning operators. Although TALplanner does use the same semantics, the extended planning language $\mathcal{L}(\text{ND})^*$ contains a new operator macro providing a syntax which facilitates the use of resource constraints and other planning-oriented concepts that are not present in standard TAL-C. This is in line with the standard TAL practice of preserving the logical base language $\mathcal{L}(\text{FL})$ and its semantics but providing different variations of the high-level macro language $\mathcal{L}(\text{ND})$ that are adapted to special tasks.

The examples below demonstrate the operator definition syntax using three of the six logistics operators. Further examples will be shown when the IPC-2002 benchmark domains are discussed.

#operator **load-truck**(*obj*, *truck*, *loc*) :at s
:precond [s] at(*obj*, *loc*) ∧ at(*truck*, *loc*)
:effects [s+1] at(*obj*, *loc*) := false, [s+1] in(*obj*, *truck*) := true

#operator **unload-truck**(*obj*, *truck*, *loc*) :at s
:precond [s] in(*obj*, *truck*) ∧ at(*truck*, *loc*)
:effects [s+1] in(*obj*, *truck*) := false, [s+1] at(*obj*, *loc*) := true

#operator **drive**(*truck*, *loc1*, *loc2*) :at s
:precond [s] at(*truck*, *loc1*) ∧ city_of(*loc1*) $\hat{=}$ city_of(*loc2*) ∧ *loc1* ≠ *loc2*
:effects [s+1] at(*truck*, *loc1*) := false, [s+1] at(*truck*, *loc2*) := true

Although not used in the simple logistics operators above, TALplanner also allows the use of context-dependent and quantified effects as well as prevail conditions. Unlike pure preconditions, prevail conditions are not limited to the invocation state of an operator but can refer to the entire interval during which the operator is executed. The interval at which each prevail condition must hold is explicitly specified, which provides additional flexibility compared to requiring that a precondition must always hold throughout the execution of an action.

### 2.5 Resources

If TALplanner was limited to generating sequential plans, resource consumption and production could be handled using plain operator effects. For example, if loading a truck requires one unit of space, the amount of available space could be decreased as follows:

#operator **load-truck**(*obj*, *truck*, *loc*) :at s
:precond [s] at(*obj*, *loc*) ∧ at(*truck*, *loc*)
:effects [s+1] space(*truck*) := value(s, space(*truck*)) − 1, ...

With concurrent planning, this is clearly not sufficient, since multiple parallel invocations of load-truck would still only consume one unit of space. For this reason, TALplanner has explicit support for resources (Kvarnström et al., 2000).

Resources can be declared in a manner similar to ordinary fluents: They can have parameters and can take values in an arbitrary integer or fixed point domain. Unlike some planners, TALplanner only provides one type of resource, but provides several types of resource effects. Resources can be produced and consumed. They can also be borrowed (and automatically returned), either exclusively, meaning that the borrower has exclusive use of the resource during the specified interval, or non-exclusively, where multiple actions can borrow the same units of a certain resource concurrently. The latter case may appear strange, but can be useful when one wants to use a resource as a semaphore or mutex. Finally, resources can be assigned a completely new value.

In the following example, loading a truck always consumes one unit of space.

#operator **load-truck**(*obj, truck, loc*) :at s
 :precond   [s] at(*obj, loc*) ∧ at(*truck, loc*)
 :effects   [s+1] at(*obj, loc*) := false, [s+1] in(*obj, truck*) := true
 :resources [s+1] :consume space(*truck*) :amount 1

Unlike ordinary fluents, a resource res has multiple aspects that can be queried and used in formulas such as operator preconditions or control rules. At any timepoint, there is an initial amount available, $init(res). A certain amount may be consumed during this time step ($consumed(res)), produced ($produced(res)), borrowed exclusively ($borrowed(res)) and borrowed non-exclusively ($borrowed-nonex(res)). This results in a remaining amount available ($available(res)), which must be between the minimum ($minimum(res)) and the maximum ($maximum(res)) allowed. The ability to refer to these aspects directly allows the user to specify more complex resource constraints than a simple minimum or maximum value for a resource, such as a control rule defining a maximum amount that may be consumed per time step.

This concludes the description of planning domain definitions in TAL. The following sections will show the structure of TALplanner's forward-chaining search tree and how the search process is constrained using control rules.

## 3. Search and Control Rules

Like any forward-chaining planner, TALplanner searches for a plan in a tree where the root corresponds to the initial state and where each outgoing edge corresponds to one of the operators applicable in its source node. Two trivial examples are shown in Figure 2, where the notation [s,t] A means that the action A is executed between time s and time t. For sequential planning (Figure 2a), a new action is always added at the time step where the previous action ended. For concurrent planning (Figure 2b), TALplanner still adds a single action at a time to the plan, but the constraint on the time where the action is executed is relaxed: The action must not start before the start of an existing action in the current plan prefix or after the end of an existing action. When searching the tree, preference is given to actions invoked at earlier timepoints. In other words, TALplanner tries to add as many applicable actions as possible at the same timepoint before stepping to the next timepoint, so in Figure 2b the subtree starting in [0,4] A3, executing the action A3 between time 0 and time 4, would have been explored before backtracking to the subtree starting with [2,5] A3, where A3 happens to take slightly less time to execute due to differences in the state where the action is invoked. The search process ends as soon as the planner has found a plan ending in a state satisfying the goal. The exact definition of the search tree is available in Kvarnström and Doherty (2000) for sequential TALplanner and Kvarnström et al. (2000) for concurrent TALplanner.

Although it is common to view each node in a search tree as consisting of a single state, and an operator as a function from states to states, this is not sufficient for TALplanner, for several reasons: A single operator may generate multiple new states, the evaluation of a temporally extended goal or domain-dependent control rule may require access to the entire state history beginning in the initial state, and during concurrent planning a future state may be modified by several operators before it reaches its final configuration. For these reasons, it is more convenient to view each node as consisting of a state sequence, or (equivalently) a logical model, as indicated in the figure.

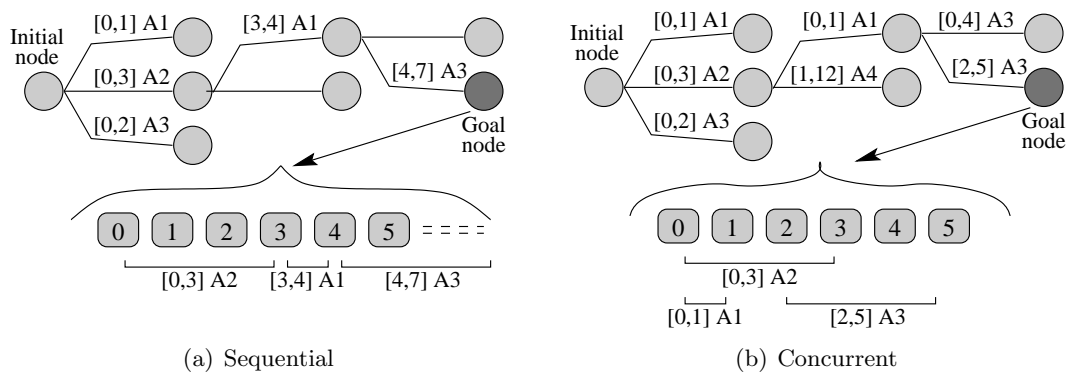

Figure 2: Forward-Chaining Search Space

A simple forward-chaining planner can be implemented by searching this tree using a standard search algorithm, such as iterative deepening or depth first search. But although using a complete search algorithm is clearly enough to make the planner complete, it is equally clear that a certain degree of goal-directedness is required to make the search process *efficient*. This is achieved using domain-dependent control rules.

### 3.1 Using Domain-Dependent Control Rules

In fully automated planning, the planner is generally only supplied with an initial state, a set of acceptable goal states (often specified using propositional or first-order formulas that must hold in any goal state), and a set of operators to be used in the plan. It is up to the planner to determine how to search for a plan efficiently, with the possible exception of various command line options that can be fine-tuned by the user.

However, in some cases the user has additional information about a planning domain that could be of use to the planner, and this information may be difficult to extract mechanically from a simple domain specification. If this is the case, it would make sense to allow the user to supply this information to the planner. Although it entails somewhat more work for the user, it may also lead to finding plans more quickly or finding plans of higher quality.

There are of course many different kinds of additional information that could be given to the planner. TALplanner (inspired by TLplan, Bacchus & Kabanza, 2000) allows the user to specify a set of first-order TAL formulas that must be entailed by the final plan. This serves two separate purposes. First, it allows the specification of complex temporally extended goals such as safety conditions that must be upheld throughout the execution of a plan, and second, the additional constraints on the final plan often allow the planner to prune entire branches of the search tree, since it can be proven that any leaf on the branch will violate at least one such goal. In many cases this pruning is the main reason for the use of the formula, in which case it is often called a control rule. (Allowing the planner to prune branches *efficiently* requires some additional analysis, as described in Kvarnström, 2002.)

### 3.2 Control Rules for the Logistics Domain

The following are some of the control rules we use for the logistics domain. Further control rule examples will be given when the IPC-2002 benchmark domains are discussed.

First, a package should only be loaded onto a plane if a plane is required to move it, i.e., if the goal requires it to be at a location in another city. Second, if we have unloaded a package from a plane, the package must have arrived in the correct city satisfying the goal. Third, if a package is at its destination, it should not be moved.

#control :name "**only-load-when-necessary**"
 [t] ¬in(*obj*, *plane*) ∧ at(*obj*, *loc*) ∧
 ¬∃*loc*' [ goal (at(*obj*, *loc*')) ∧ [t] city_of(*loc*) $\not\hat{=}$ city_of(*loc*') ] →
 [t+1] ¬in(*obj*, *plane*)

#control :name "**only-unload-when-necessary**"
 [t] in(*obj*, *plane*) ∧ at(*plane*, *loc*) ∧
 ¬∃*loc*' [ goal (at(*obj*, *loc*')) ∧ [t] city_of(*loc*) $\hat{=}$ city_of(*loc*') ] →
 [t+1] in(*obj*, *plane*)

#control :name "**objects-remain-at-destinations**"
 [t] at(*obj*, *loc*) ∧ goal (at(*obj*, *loc*)) → [t+1] at(*obj*, *loc*)

Note that these rules could of course be expressed on various other logically equivalent forms. Most such variations would have identical performance, since TALplanner internally normalizes many aspects of control formulas during its domain analysis phase.

## 4. The Third International Planning Competition

In the second international planning competition (IPC-2000), the planning domains used mainly STRIPS expressivity. Support for typed objects was not required, and for those domains that could use ADL-style quantified and conditional effects, restricted STRIPS versions were also provided.

Although we did expect some increase in expressivity in the third competition (IPC-2002), we were quite surprised by the extent of the changes. Fortunately, TALplanner already supported many of the new requirements, and some of the others were easily implemented. Despite this we did make some rather significant changes in order to handle the *combination* of all these extensions more efficiently. Below we will discuss how each of the new requirements affected TALplanner together with a few other improvements that have been prompted by the domains used in the competition.

### 4.1 ADL-style Operator Definitions

Though there were STRIPS versions of most planning domains in IPC-2002, the more complex versions of the domains required the use of quantified conditional effects. Like most other current planners, TALplanner is not limited to STRIPS expressivity and already had support for this.

### 4.2 Numeric Types and Arithmetic

All IPC-2000 domains that required numeric values emulated these values using ordinary objects. In the Miconic-10 elevator domain, for example, floor numbers were emulated using objects named f0, f1, and so on. The next floor was not calculated as $f + 1$ but by using an explicitly defined predicate above(floor, floor).

The same approach was taken in the simplest versions of the IPC-2002 domains, but there were also "Numeric" versions of these domains where numeric types were required and where arithmetic operators were used. This was already supported by TALplanner, but unfortunately there was not enough time to write control rules for these domains.

### 4.3 Concurrency

Despite the fact that some IPC-2000 domains provided the potential for using concurrent actions, such as driving several trucks concurrently in the logistics domain, there was no reward for exploiting this potential. Plan quality was measured in terms of the number of operators in a plan, not in terms of the amount of time required to execute the plan. Consequently, several planners (including TALplanner) only generated sequential plans, even for highly concurrent domains.

In IPC-2002, plan quality was mainly measured in terms of the timepoint at which the last operator finished executing (the "makespan" of the plan, in scheduling terms), and any planner generating sequential plans would have been severely handicapped. Fortunately a concurrent version of TALplanner had already been implemented, together with support for resources (Kvarnström et al., 2000), and could be used in the competition.

Although concurrent TALplanner had already been applied to a number of domains, the competition provided us with a more varied set of domains that sometimes exploited concurrency in slightly different ways. This provided us with new ideas for improvements to TALplanner, and several minor enhancements to TALplanner's formula analysis algorithms were implemented during the first phase of the competition, allowing it to handle certain types of control formulas more efficiently when doing concurrent planning.

### 4.4 Operators with Non-Unit and Context-Dependent Duration

In IPC-2000, each plan operator used a single time step. In the SimpleTime and Timed versions of the IPC-2002 planning domains, operators could have a non-unit duration, so that (for example) walking requires more time than driving. This was already supported by TALplanner, and no changes were required.

In the Timed versions of the IPC-2002 planning domains, the durations of some operators could also be context-dependent, and could be specified using arithmetic expressions, requiring support for numeric types as already discussed above. For example, the time required to drive a truck between two locations could be specified as the distance between the locations divided by the speed of this particular truck. This was also already supported by TALplanner.

TALplanner also permits effects to take place at multiple timepoints within the duration of an action, although this was not used in the competition.

### 4.5 Non-Integer Time

Some of the IPC-2002 contest domains required operator durations to be calculated with a precision of at least three decimals, which posed a problem for us. The underlying TAL-C logic is based on integer time, and therefore the same is true for TALplanner. Introducing non-integer time properly would have required changes to the underlying TAL semantics, which could not be done in the time that was available, and therefore we simply multiplied durations by a thousand. When printing a plan, all time values were divided by a thousand.

### 4.6 Operators with Extended Duration

In the initial implementation of TALplanner (in 1998–1999), it was assumed that although operators might have extended durations, something interesting would be happening at a significant proportion of the discrete time steps within that duration. For example, an operator invoked at $t$ might have a duration of 5 time steps, where some effects take place at time $t+1$, some at time $t+4$, and some at time $t+5$. This assumption influenced some of the algorithms and data structures in TALplanner, and appeared reasonable at the time, since most planning domains in the literature only used single-step operators.

Nevertheless, it was always our intention to extend these algorithms and structures to handling plans with sparse effects, where most discrete time steps contain no effects at all. Doing this would not have been difficult, but partly for that very reason – there were more interesting research issues to be tackled instead – it was continuously postponed.

IPC-2002 finally provided us with a compelling reason to change the data structures, together with a number of example domains that could be used to test the changes. For example, an operator in a timed domain from IPC-2002 might have a duration of (say) 89.237, requiring 89237 discrete time steps, where all effects take place at the beginning or at the end of the action. This led us to implement a new sparse state structure and change a few algorithms whose time complexity accidentally depended on the duration of an operator rather than the number of time steps where something actually happened. The current version of TALplanner allows both state structures to be used depending on the characteristics of each planning domain.

### 4.7 The "No Moving Targets" Rule

As already mentioned, TALplanner's semantics is based on the use of TAL, while the planning competition uses PDDL2.1. While the semantic differences between these two approaches are usually not a major problem, we did have some trouble with the way the effects of durative actions are modeled in PDDL2.1. In essence, PDDL2.1 predicates or numerical fluents that are affected by the effects of an action are considered to be "moving targets", and the preconditions of another action are not allowed to refer to them at the same timepoint. Instead, a certain intermediate interval (arbitrarily chosen to be 0.001 units of time) is required between the assertion of a fact and the subsequent use of that fact, even at the beginning of the plan where actions cannot begin exactly at time 0. In TAL, effects taking place at time $t$ are assumed to give fluents their new values exactly at that timepoint, and those values can immediately be used. If there is some uncertainty in the exact time when the effect takes place, one can for example explicitly state that the value is unknown during the inner part of a certain interval but is known at the end of that interval (though this is not yet implemented in TALplanner).

Changing TALplanner to use the exact PDDL2.1 semantics was out of the question, since this would change some of the most fundamental assumptions in the planner. Instead it was necessary to come up with a workaround that let us simulate this semantics. There are several ways this could be done. One method would involve making minor changes to the action definitions in order to assert the final effects of each action slightly later (0.001 units of time later, to be exact). During the competition we instead implemented a trivial modification to the way a plan is printed: At any timepoint where something happens in the plan (for example, where an operator is invoked), an additional delay of 0.001 is inserted. This ensures that all plans are safe according to PDDL2.1 semantics but sometimes leads to generating slightly worse plans than necessary.

### 4.8 Finding Shortest Paths

In the Rover and DriverLog domains, vehicles and/or people must travel along road networks, where different roads may have different costs (lengths) and where it is essential to take the shortest path between any two points.

Although it is possible to define a shortest path algorithm using TALplanner's input language, the formulas become somewhat complicated. Finding the shortest path between two locations in a weighted graph of places and roads seems to be useful in many domains, and therefore such an algorithm was implemented directly in the planner.

In fact, two algorithms were implemented: One for finding the cost of the shortest path between two given locations, and one for finding the distance to the closest location satisfying a given formula (for example the closest location which is a reasonable destination for a certain truck in the DriverLog domain). These functions can be called from control rules in order to ensure that each step one takes leads to a location which is on some shortest path to the current destination.

## 5. Modeling the Competition Domains

Of the eight planning domains in the third International Planning Competition, six were intended for hand-tailored planners. Except for the final domain, UMTranslog-2, all domains exist in at least four different variations: STRIPS, Numeric (where numeric quantities are involved), SimpleTime (where operators take constant non-unit time), and Timed (where operator durations may depend on the actual parameters in a specific operator invocation). TALplanner participated in all six domains, but due to lack of time for creating control rules, we limited our participation to the STRIPS, SimpleTime, and Timed versions of the domains.

In this section we will describe how the domains were translated from PDDL2.1 to TALplanner, and discuss some of the control rules that were created to handle the domains more efficiently. The main focus will be on two domains: ZenoTravel and Satellite. For these domains we will describe most of the control rules that were used in the competition as well as the incremental process of creating the rules, omitting only a few technical details and a couple of complex rules that turned out to have minimal impact on planner performance and plan quality. For the remaining domains (Depots, DriverLog, Rovers, and UMTranslog-2) we will describe the general intuitions behind our control rules, omitting the actual formulas due to space restrictions. First, though, we will begin with a few comments on the process of formalizing planning domains.

### 5.1 Using Pre-defined PDDL Domains: Half the Work in Twice the Time?

In order to create a formal description of a real-world planning domain, it is of course always necessary to have a thorough understanding both of the domain itself and of how plans for the domain are eventually going to be used. There are several reasons why this is required, and most of these reasons are equally valid regardless of whether the formalization will eventually be used as the input to a fully automated planner or to a hand-tailored planner like TALplanner.

First, understanding the domain is required in order to determine what aspects of the domain truly need to be modeled (as types, predicates and functions) and what aspects can be abstracted away. For example, the standard formalization of the logistics domain does not model distances between locations, but allows trucks to move between any two locations in one time step. This is sufficient for some purposes, but a plan that is optimal given this abstraction may be extremely suboptimal if actually carried out by real trucks, which usually lack teleportation abilities. Similarly, it does not model package sizes or weights, or cargo capacities for trucks or airplanes. Neither does it model truck drivers, acceptable working hours for drivers, the additional costs incurred by overtime pay, or time required for maintenance activities such as changing to winter tires once a year. Which of these aspects need to be modeled depends very much on the particular application one has in mind.

Second, a detailed understanding of the domain is required in order to determine what operators are available to the planner and exactly how their preconditions and effects should be represented within the abstract logical model of the domain.

And finally, for hand-tailored planners, the domain must be understood in order to be able to guide a search algorithm using domain-dependent heuristics or control rules.

Usually all of these aspects of a domain are modeled at the same time, and much of the information and knowledge about the domain that was gathered in order to find a suitable set of predicates and operators – which is needed even for a fully automated planner – can be reused in the development of control rules or heuristics for a hand-tailored planner.

In the planning competition, however, the task is divided into two parts: The organizers define a set of domains using PDDL2.1, and then it is up to the competitors in the hand-tailored track to find suitable ways of guiding their planners. In one way, one could say that the competitors only need to do half the work, since the formalization is already done and only the task of finding control rules remains. Unfortunately it is still necessary to understand the domain just as thoroughly in order to write control rules. For the more complex domains, doing this half of the work in isolation might easily take twice the time, since all the constraints involved in the domain have to be understood from a PDDL2.1 formalization rather than by talking to domain experts. This is especially true for the complex UMTranslog-2 logistics domain, where a significant amount of time was spent trying to determine exactly how packages were allowed to move and how they can be loaded into and unloaded from various kinds of vehicle.

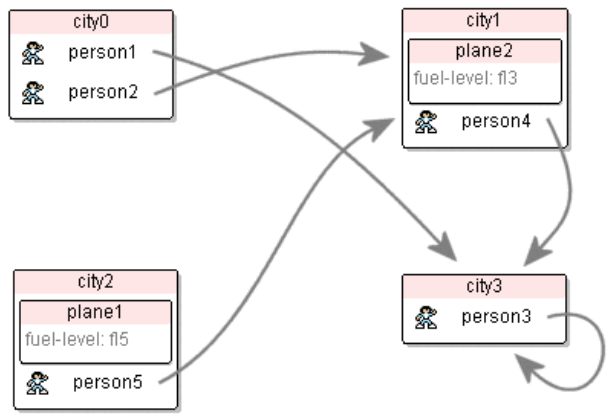

Figure 3: A ZenoTravel problem instance (STRIPS problem 6)

Another problem caused by having to use a predefined formalization of a planning domain is that the degree of detail used in the model is determined in advance. In the real world there would more likely be a *minimum* level of detail required, and anything above this level would be acceptable. It may not seem like this should be a problem – intuitively, adding new details to a planning problem ought to make it harder, and so it would be best to remain at the minimum level of detail. But this is not always true, especially not when control rules are involved. This will be seen in the timed ZenoTravel domain, for example, where some control rules would be both simpler and more effective if it was possible to refuel to a specific level, just like in the real world, rather than just having a simple abstract refuel operator that unconditionally fills the tank completely.

This should not be taken as a complaint against the organization of the competition – allowing different planners to use different formalizations would of course be completely infeasible. Nevertheless, it does present some additional problems that are not encountered to the same degree in real-world domains and that deserve to be mentioned here.

## 5.2 The ZenoTravel Domain

In the ZenoTravel domain, there are a number of aircraft that can fly people between cities. There are five actions available: Persons may board and debark aircraft, and aircraft may fly, zoom (fly quickly, using more fuel), and refuel. There are no restrictions on how many people an aircraft can carry. Flying and zooming are equivalent except that zooming is generally faster and uses more fuel. Figure 3 shows an example problem, with arrows pointing out goal locations.

### 5.2.1 ZenoTravel: STRIPS

Below we show the operator definitions for the STRIPS version of the ZenoTravel domain. These operators have been more or less directly translated from the PDDL representation. The main difference is that the PDDL representation uses PDDL2.1 level 1, with single-step actions, which has a stricter concept of mutual exclusion than TALplanner does and

Kvarnström & Magnusson

automatically enforces certain invariants, such as the fact that an aircraft should not leave if a person is boarding, because the location of the aircraft is modified by fly and used in the precondition of board. The TAL-C semantics used by TALplanner is more similar to PDDL2.1 level 3 (with durative actions), where such invariant conditions must be stated explicitly. This is done using prevail conditions, which are considered to be separate from true *pre*-conditions. Note that in the STRIPS formalization fly and zoom take the same amount of time, since only single-step actions are possible.

#operator **board**(*person*, *aircraft*, *city*) :at t
 :precond   [t] at(*person*, *city*) ∧ at(*aircraft*, *city*)
 :prevail   [t+1] at(*aircraft*, *city*)
 :effects   [t+1] at(*person*, *city*) := false, [t+1] in(*person*, *aircraft*) := true

#operator **debark**(*person*, *aircraft*, *city*) :at t
 :precond   [t] in(*person*, *aircraft*) ∧ at(*aircraft*, *city*)
 :prevail   [t+1] at(*aircraft*, *city*)
 :effects   [t+1] in(*person*, *aircraft*) := false, [t+1] at(*person*, *city*) := true

#operator **fly**(*aircraft*, *city1*, *city2*, *flevel1*, *flevel2*) :at t
 :precond   [t] at(*aircraft*, *city1*) ∧ fuel-level(*aircraft*, *flevel1*) ∧ next(*flevel2*, *flevel1*)
 :effects   [t+1] at(*aircraft*, *city1*) := false, [t+1] fuel-level(*aircraft*, *flevel1*) := false,
            [t+1] at(*aircraft*, *city2*) := true, [t+1] fuel-level(*aircraft*, *flevel2*) := true

#operator **zoom**(*aircraft*, *city1*, *city2*, *flevel1*, *flevel2*, *flevel3*) :at t
 :precond   [t] at(*aircraft*, *city1*) ∧ fuel-level(*aircraft*, *flevel1*) ∧
              next(*flevel2*, *flevel1*) ∧ next(*flevel3*, *flevel2*)
 :effects   [t+1] at(*aircraft*, *city1*) := false, [t+1] fuel-level(*aircraft*, *flevel1*) := false,
            [t+1] at(*aircraft*, *city2*) := true, [t+1] fuel-level(*aircraft*, *flevel3*) := true

#operator **refuel**(*aircraft*, *city*, *flevel*, *flevel1*) :at t
 :precond   [t] fuel-level(*aircraft*, *flevel*) ∧ next(*flevel*, *flevel1*) ∧ at(*aircraft*, *city*)
 :prevail   [t+1] at(*aircraft*, *city*)
 :effects   [t+1] fuel-level(*aircraft*, *flevel*) := false, [t+1] fuel-level(*aircraft*, *flevel1*) := true

After translating the operator definitions, it is time to create a set of control rules. There are basically two ways of doing this: First, one can sit down and think about suitable properties for a plan, and then write control rules that ensure that these properties will hold. Second, one can instruct the planner to show each branch that is explored in the search tree, and by observing the output one can identify "obviously stupid" choices made by the planner, such as choosing an action instance that inevitably leads to backtracking or performing actions that are useless given the goals. Control rules can then be written to prevent these branches of the tree from being explored. Both of these approaches will be covered here.

We begin with the first method, attempting to find a number of reasonable control rules simply by thinking about the properties of the ZenoTravel domain. Given some experience from other planning domains, this is in fact quite easy. For example, in many domains there are certain goals such that once they are satisfied, one should never allow them to be destroyed. In the ZenoTravel domain, people who are at their destinations never need to board an aircraft, which gives rise to the following control rule:

#control :name "**only-board-when-necessary**"
    [t] ¬in(*person*, *aircraft*) ∧ [t+1] in(*person*, *aircraft*) →
    ∃*city*, *city2* [ [t] at(*person*, *city*) ∧ goal(at(*person*, *city2*)) ∧ *city* ≠ *city2* ]

This TAL formula states that if we have a state transition from the person not being in the aircraft at time $t$ to the person being in the aircraft at time $t+1$, (that is, if the person just boarded the aircraft), then there must be a reason why this is allowed: The person must be in a certain city and there must be a goal that the person should be in another city.

As noted previously control formulas can usually be written in many different forms. For example, it would have been equally valid to state that if a person is at a city (and therefore not in an aircraft), and is not required to be somewhere else, then at the next timepoint that person should still not be on board an aircraft:

#control :name "**only-board-when-necessary**"
    [t] at(*person*, *city*) ∧ ¬∃*city2* [ goal(at(*person*, *city2*)) ∧ *city* ≠ *city2* ] →
    [t+1] ¬in(*person*, *aircraft*)

Note that although it may at first glance appear that a planner would have to be extraordinarily stupid to destroy goals that have already been satisfied, there are also many cases where temporarily destroying a goal is necessary in order to satisfy other goals. For example, if there is a goal that a certain aircraft should be at a certain location and it has already reached that destination, it might still have to fly a number of people to their destinations before it can return to its own destination.

Another natural idea (since aircraft do not follow predetermined routes in ZenoTravel, as they usually do in real life) would be to say that people should only debark when they have reached their final destination:

#control :name "**only-debark-when-in-goal-city**"
    [t] in(*person*, *aircraft*) ∧ [t+1] ¬in(*person*, *aircraft*) →
    ∃*city* [ [t] at(*aircraft*, *city*) ∧ goal(at(*person*, *city*)) ]

There is a potential problem with this rule: In some cases an optimal plan might require a number of people to debark one plane and then board a number of other planes, which could fly them to their destination concurrently, and this is strictly forbidden by only-debark-when-in-goal-city. This is a common problem that occurs for many planning domains, and it is up to the user to determine what to do depending on the requirements of the application for which the planner is being used.

There are a number of possible choices: We could ignore this problem and accept suboptimal plans, skip the rule completely and let the planner search through a vastly greater search space in order to find a plan which is guaranteed to be optimal, or as a compromise, attempt to create a weaker rule that does cut down the search space to some degree but gives optimal or closer-to-optimal plans. During the planning competition the conditions were somewhat artificial and were not clearly stated – would it be beneficial for a planner to spend ten times as much effort finding a plan if this plan was only five percent better, on average? We guessed that this would not be the case, and consequently we chose to include the control rule as stated above.

In the future, a better solution would most likely be to *prefer* those plans where a person does not debark before reaching his destination but still *allow* other plans. This alternative will be discussed in more detail in the conclusions.

Given these two rules, we might now continue with the second approach to finding control rules. We run TALplanner on a simple problem instance and consider the operator sequences the planner examines during the depth-first search process. This is the beginning of such a sequence for the problem instance in Figure 3. The complete plan generated by the planner contains 123 operators and requires 60 time steps. It is shown here in the IPC-2002 STRIPS result format where the timepoint at which an action is invoked is followed by the action instance.

```
0: (board person4 plane2 city1)      4: (debark person2 plane2 city1)    8: (refuel plane1 city0 fl0 fl1)
0: (board person5 plane1 city2)      4: (debark person5 plane1 city1)    8: (refuel plane2 city1 fl0 fl1)
1: (fly plane1 city2 city0 fl5 fl4)  5: (fly plane1 city1 city0 fl3 fl2) 9: (fly plane1 city0 city1 fl1 fl0)
1: (fly plane2 city1 city0 fl3 fl2)  5: (fly plane2 city1 city0 fl1 fl0) 9: (fly plane2 city1 city0 fl1 fl0)
2: (board person1 plane1 city0)      6: (fly plane1 city0 city1 fl2 fl1) 10: (refuel plane1 city1 fl0 fl1)
2: (board person2 plane2 city0)      6: (refuel plane2 city0 fl0 fl1)    11: (fly plane1 city1 city0 fl1 fl0)
3: (fly plane1 city0 city1 fl4 fl3)  7: (fly plane1 city1 city0 fl1 fl0) 11 : (refuel plane2 city0 fl0 fl1)
3: (fly plane2 city0 city1 fl2 fl1)  7: (fly plane2 city0 city1 fl1 fl0) ...
```

The beginning of the operator sequence appears to be reasonable, but after time 4, airplanes seem to be flying around randomly. There are no control rules guiding them, so apparently it was mainly luck that caused the planes to find reasonable cities to fly to at time 1 and 3. To make airplanes more goal-directed, we identify three important reasons why an airplane should move from city to city2: that the goal asserts that the aircraft must end up in city2 when the plan is complete, that one of its passengers wants to go to city2, or that there is a person waiting to be picked up by an airplane in city2. The following rule formalizes these three intuitions:

#control :name "**planes-always-fly-to-goal**"
    [t] at(*aircraft*, *city*) ∧ [t+1] ¬at(*aircraft*, *city*) →
    ∃*city2* [ [t+1] at(*aircraft*, *city2*) ∧
        (goal(at(*aircraft*, *city2*)) ∨
        ∃*person* [ [t] in(*person*, *aircraft*) ∧ goal(at(*person*, *city2*)) ] ∨
        ∃*person* [ [t] at(*person*, *city2*) ∧ goal(¬at(*person*, *city2*)) ]) ]

With these control rules, TALplanner can quickly produce a set of plans for the 20 "hand-coded" problems from the IPC-2002 competition, and although the plans will not be optimal, they will not be nearly as bad as the example given above. Together, the plans require a total of 7164 operators and 618 time steps. The plan for the example in Figure 3 requires 20 operators and 7 time steps.

Nevertheless, there are still some improvements that can be made. The first criterion is too admissible: It allows a plane to visit its destination even if it still needs to pick up or drop off passengers. One way of preventing this would be to add the condition that all passengers must have reached their destinations:

#define [t] **all-persons-arrived**:
    ∀*person*, *city* [ goal(at(*person*, *city*)) → [t] at(*person*, *city*) ]

#control :name "**planes-always-fly-to-goal**"
  [t] at(*aircraft*, *city*) ∧ [t+1] ¬at(*aircraft*, *city*) →
  ∃*city2* [ [t+1] at(*aircraft*, *city2*) ∧
         (**[t] all-persons-arrived** ∧ goal(at(*aircraft*, *city2*)) ∨
          ∃*person* [ [t] in(*person*, *aircraft*) ∧ goal(at(*person*, *city2*)) ] ∨
          ∃*person* [ [t] at(*person*, *city2*) ∧ goal(¬at(*person*, *city2*)) ]) ]

This improves plan quality slightly, and TALplanner now requires 7006 operators and 575 time steps. But the new control rule is in fact too strict, which can be seen in the following plan tail for handcoded STRIPS problem number 3:

14: (fly plane2 city4 city7 fl2 fl1)
14: (fly plane4 city8 city9 fl3 fl2)
14: (refuel plane1 city6 fl2 fl3)
14: (refuel plane3 city9 fl4 fl5)
15: (debark person24 plane4 city9)
15: (debark person28 plane4 city9)
15: (debark person34 plane2 city7)
15: (refuel plane1 city6 fl3 fl4)
15: (refuel plane2 city7 fl1 fl2)
15: (refuel plane3 city9 fl5 fl6)
15: (refuel plane4 city9 fl2 fl3)
16: (fly plane1 city6 city8 fl4 fl3)
16: (fly plane3 city9 city4 fl6 fl5)

In this example, plane1 and plane3 had to wait until all passengers had debarked from several other planes until they could go to their final destinations, even though we can clearly see that there was no real reason for them to wait, because all potential passengers had already been picked up and plane1 and plane3 already had enough fuel. We once again alter the control rule according to this new insight: A plane can go to its final destination if all passengers on board the plane are headed towards the same destination and there is no person left to be picked up (that is, all persons have already arrived or are currently on board planes).

#define [t] **all-persons-arrived-or-in-planes**:
  ∀*person*, *city* [ goal(at(*person*, *city*)) → [t] at(*person*, *city*) ∨ ∃*aircraft* [ in(*person*, *aircraft*) ] ]

#control :name "**planes-always-fly-to-goal**"
  [t] at(*aircraft*, *city*) ∧ [t+1] ¬at(*aircraft*, *city*) →
       [t+1] at(*aircraft*, *city2*) ∧
       ((goal(at(*aircraft*, *city2*)) ∧ **[t] all-persons-arrived-or-in-planes** ∧
         **∀person [ [t] in(person, aircraft) → goal(at(person, city2)) ]**) ∨
        ∃*person* [ [t] in(*person*, *aircraft*) ∧ goal(at(*person*, *city2*)) ] ∨
        ∃*person* [ [t] at(*person*, *city2*) ∧ goal(¬at(*person*, *city2*)) ])]

This yields another minor improvement, and TALplanner now requires 6918 operators and 564 time steps. For the example used above, the end of the plan now looks as follows:

14: (fly plane1 city6 city8 fl2 fl1)
14: (fly plane2 city4 city7 fl2 fl1)
14: (fly plane4 city8 city9 fl3 fl2)
14: (refuel plane3 city9 fl4 fl5)
15: (debark person24 plane4 city9)
15: (debark person28 plane4 city9)
15: (debark person34 plane2 city7)
15: (fly plane3 city9 city4 fl5 fl4)

We once more study the plans generated by the current set of rules and quickly identify another obvious problem: Any number of airplanes may fly to the same location to pick up the same person. Once again, it is necessary to find a reasonable balance between finding optimal plans and finding plans quickly. In the contest, we attempted to find a high quality (but probably non-optimal) plan as quickly as possible. This was done by ensuring that no more than one airplane may go to any given place at the same time, *if* the sole purpose for going there is to pick up a person who is waiting:

#control :name "**planes-always-fly-to-goal**"
   [t] at(*aircraft*, *city*) ∧ [t+1] ¬at(*aircraft*, *city*) →
   ∃*city2* [ [t+1] at(*aircraft*, *city2*) ∧
          ((goal(at(*aircraft*, *city2*)) ∧ [t] all-persons-arrived-or-in-planes ∧
            ∀*person* [ [t] in(*person*, *aircraft*) → goal(at(*person*, *city2*)) ]) ∨
            ∃*person* [ [t] in(*person*, *aircraft*) ∧ goal(at(*person*, *city2*)) ] ∨
            ∃*person* [ [t] at(*person*, *city2*) ∧ goal(¬at(*person*, *city2*)) ] ∧
              ¬∃**aircraft2** [ **[t+1] at(aircraft2, city2)** ∧ **aircraft2 ≠ aircraft** ])]

This rule provides a major improvement, and the complete set of plans now requires 5075 operators and 434 time steps.

So far, we have controlled where airplanes fly, when people board an airplane, and when they debark. There are no rules governing refueling, and a quick look at a plan for one of the larger problem instances reveals that whenever an aircraft has nothing else to do, it will refuel. This seems a little bit wasteful, but we are satisfied with adding a rule stating that airplanes must only refuel when their tanks are empty. This rule is not perfect, since an airplane may miss an opportunity to "pre-emptively" refuel and it can still refuel one fuel level even if it is not going to fly, but it does provide a significant improvement, bringing the number of operators down to 4234. The number of time steps is still 434.

A few minor adjustments were made to these rules before they were used in the competition. These adjustments include a modification to only-board-when-necessary to ensure that a person who must travel from *city* to *city2* will choose a plane that already needs to visit both *city* and *city2*, if this is possible, since this is less likely to increase the total number of flights.

One final change is prompted by the fact that the intended differences in timing between fly and zoom cannot be modelled correctly in the STRIPS version of the domain. Since all operators must take the same amount of time, the only difference between these two operators is that zoom uses twice as much fuel. Although it would have been possible to add a control rule ensuring that zoom was not used, it was easier to simply remove the zoom operator from the domain definition.

5.2.2 ZenoTravel: SimpleTime

The SimpleTime version of ZenoTravel is quite similar to the STRIPS version, the only difference being that actions may have non-unit duration and that certain preconditions must hold throughout the execution of an action. The TALplanner operator definitions are changed accordingly. For example, the board and fly operators can be changed as follows:

#operator **board**(*person*, *aircraft*, *city*) :at t
 :precond    [t] at(*person*, *city*) ∧ at(*aircraft*, *city*)
 :prevail    **[t+1, t+20]** at(*aircraft*, *city*)
 **:duration  20**
 :effects    [t+1] at(*person*, *city*) := false, **[t+20]** in(*person*, *aircraft*) := true

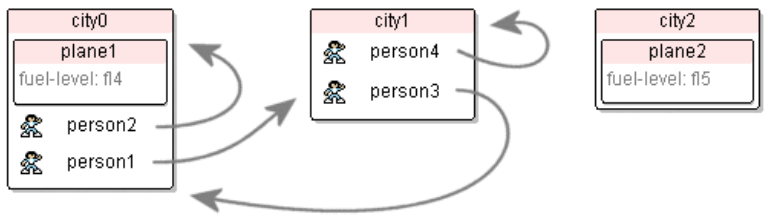

Figure 4: A ZenoTravel problem instance (SimpleTime problem 3)

#operator **fly**(*aircraft*, *city1*, *city2*, *flevel1*, *flevel2*) :at t
:precond   [t] at(*aircraft*, *city1*) ∧ fuel-level(*aircraft*, *flevel1*) ∧ next(*flevel2*, *flevel1*)
**:duration  180**
:effects   [t+1] at(*aircraft*, *city1*) := false, [t+1] fuel-level(*aircraft*, *flevel1*) := false,
           **[t+180]** at(*aircraft*, *city2*) := true, **[t+180]** fuel-level(*aircraft*, *flevel2*) := true

If we run the planner on a set of SimpleTime problem instances, we get almost immediate results: The planner claims that there is no plan for any of the instances. The reason for this is, of course, that the control rules must be satisfied in any valid plan, and those rules were designed with the underlying assumption that actions had unit duration. For example, consider planes-always-fly-to-goal, which states that if a plane leaves a city at time t, it should be at a meaningful destination at t+1. When the fly action is invoked the plane must be at some city *city1*, but beginning at the next time step there will be an interval where the aircraft is not present in any city at all, until it finally arrives in *city2* 180 time steps later. In other words, planes-always-fly-to-goal now ensures that the fly operator cannot be used at all, which is not quite what was originally intended.

One way of solving this problem would be to alter planes-always-fly-to-goal to say that if a plane leaves a city at time t, it should be at a meaningful destination at t+180. Unfortunately, the duration of the flight would then be encoded directly in the control rule instead of only in the operator, and so it would not work in the Timed version, where operators have variable durations – in fact, it would not even work in SimpleTime, because the zoom operator must also be taken into account.

Instead, the domain model is augmented with a new fluent flying-to(aircraft, city) which keeps track of whether a plane is flying, and if so, what its destination is. To ensure that this fluent is kept up-to-date, the following is added to the effects of the fly and zoom operators:

[t+1] flying-to(*aircraft*, *city2*) := true, [t+180] flying-to(*aircraft*, *city2*) := false  // **for fly**
[t+1] flying-to(*aircraft*, *city2*) := true, [t+100] flying-to(*aircraft*, *city2*) := false  // **for zoom**

The planes-always-fly-to-goal rule above can now be changed as follows, stating that if an aircraft ceases to be at *city*, then it must be flying to a reasonable destination:

#control :name "**planes-always-fly-to-goal**"
   [t] at(*aircraft*, *city*) ∧ [t+1] ¬at(*aircraft*, *city*) →
   ∃*city2* [ [t+1] **flying-to**(*aircraft*, *city2*) ∧ . . . ]

The same problem arises for boarding, and a new fluent boarding(person, aircraft) is added and used whenever necessary. Given these changes, the following are the first steps of the plan generated by TALplanner for the problem instance in Figure 4, shown in the IPC-2002 timed result format where the timepoint at which an action is invoked is followed by the action instance and the duration of the action:

0: (board person1 plane1 city0) [20]
20: (fly plane1 city0 city1 fl4 fl3) [180]
20: (zoom plane1 city0 city1 fl4 fl3 fl2) [100]

Intuitively, flying and zooming plane1 at the same time should be impossible, but we have forgotten to specify this to the planner. Both actions have their preconditions satisfied at time 20, there are no prevail conditions, and the effects of the actions do not contradict each other since they take place at different timepoints: fly ends at time 200, while zoom ends at time 120.

There are several ways of specifying that fly and zoom are mutually exclusive. For example, it would be possible to introduce an interval effect stating that flying-to(*aircraft*, *city2*) must hold throughout the inner execution intervals of these actions, and become false at the end of each action:

[t+1,t+179] flying-to(*aircraft*, *city2*) := true, [t+180] flying-to(*aircraft*, *city2*) := false // **for fly**
[t+1,t+ 99] flying-to(*aircraft*, *city2*) := true, [t+100] flying-to(*aircraft*, *city2*) := false // **zoom**

It would also be possible to use a semaphore resource: An aircraft-specific resource with an initial value of 1, which can be borrowed exclusively by the fly and zoom actions. When one of these solutions is used, TALplanner finally rewards us with a short and correct plan:

0: (board person1 plane1 city0) [20]
20: (fly plane1 city0 city1 fl4 fl3) [180]
200: (board person3 plane1 city1) [20]
200: (debark person1 plane1 city1) [30]
230: (fly plane1 city1 city0 fl3 fl2) [180]
410: (debark person3 plane1 city0) [30]
;; Plan length 6, maxtime 440

Can it be improved? Remember that the STRIPS version never made use of the zoom operator. But in the SimpleTime version, flying takes 180 time steps and uses one unit of fuel, zooming takes 100 time steps and uses two units of fuel, and refueling one unit takes 73 time steps. $180 + 73$ is more than $100 + 2 \cdot 73$ and therefore we have the opposite situation: zoom is always better than fly. Commenting out the unwanted fly operator yields the following plan:

0: (board person1 plane1 city0) [20]
20: (zoom plane1 city0 city1 fl4 fl3 fl2) [100]
120: (board person3 plane1 city1) [20]
120: (debark person1 plane1 city1) [30]
150: (zoom plane1 city1 city0 fl2 fl1 fl0) [100]
250: (debark person3 plane1 city0) [30]
;; Plan length 6, maxtime 280

### 5.2.3 ZenoTravel: Timed

The Timed version further complicates the timing of the actions. Boarding and disembarking times are constant but problem-specific and are defined in the respective problem definition as two new functions, boarding-time and debarking-time. Refueling always fills the plane to its maximum capacity, but consumes time relative to the amount of fuel received and the refuel-rate of the aircraft. Each aircraft also has a fast-speed and a slow-speed with corresponding fast-burn and slow-burn fuel consumption. The distances between cities are specified using the distance(*city1*, *city2*) function.

In the Timed version, operator durations have to be correctly calculated with a precision of three decimals, prompting the TALplanner changes discussed in Sections 4.5 and 4.6. Once these extensions to TALplanner had been implemented, few changes were needed to transform the SimpleTime domain to the Timed version.

The most important difference was perhaps the fact that depending on the speed and fuel consumption values defined in each problem and the situation where the operator is used, it is sometimes better to use the fly operator and sometimes better to use the zoom operator, unlike the STRIPS version where fly was always better and the SimpleTime domain where zoom was always better.

So when is zooming better than flying? It may seem like it would be easy to answer this question, given that we are only interested in minimizing time: Just check whether refueling the aircraft sufficiently to be able to zoom, followed by zooming to the destination, would be faster than only refueling enough to be able to fly and then flying more slowly to the destination. This is handled by the first clause in use-fly-instead-of-zoom below. The precondition of fly is then altered to require that use-fly-instead-of-zoom be true, and the precondition of zoom requires that use-fly-instead-of-zoom be false. If we had been interested in minimizing a combination of time and fuel usage, then this could also have been taken into account.

This is not quite sufficient to handle all problems, though. An airplane has a maximum fuel capacity, so if its destination is too distant, it may not be able to zoom. This is handled by the second clause in use-fly-instead-of-zoom.

Yet another problem is that it is not possible to tie one refueling action to each flight, as one would expect in the real world. There are two reasons for this problem.

First, airplanes may already have some fuel in the initial state, so in some situations a plane might zoom to its destination without incurring any additional cost, again assuming that the time required for executing the plan is the only metric being used – the plane already had enough fuel anyway and never had to refuel.

Second, unlike the SimpleTime version, an airplane cannot refuel "just enough" – the refuel operator always fills the tank completely. This change was most likely introduced in order to make the planning task easier by reducing the number of possible actions to choose from (for example, a planner that needs to create all ground instances of each operator might have some trouble if the refuel operator would take the amount of fuel as a floating point argument). But despite the probable intention behind this change, it introduces new problems for our control formulas. If a plane's tank is half full and this is enough fuel to zoom from A to B, it might then have to fill the *entire* tank before continuing to C, while if it used the fly operator, it might be able to continue to C without refueling at all. This means that one would have to take all possible future flights into account when determining whether to fly or zoom. If the domain had been modeled in more detail, this problem would not have existed.

Given these two complications, guaranteeing an optimal or near-optimal plan using a control rule is not easy, which is indeed only to be expected. For the competition we decided to be satisfied with a heuristic compromise, adding a third clause to use-fly-instead-of-zoom ensuring that if zooming would require refueling *immediately* but flying would not, the fly operator would be used.

// Fly is (probably) better than zoom if:
#define [t] **use-fly-instead-of-zoom(aircraft, city1, city2)**:
　　// If fly is faster wrt speed and refueling.
　　([t] (10000 / slow-speed($aircraft$) + 10000 * slow-burn($aircraft$) / refuel-rate($aircraft$)) <
　　　　(10000 / fast-speed($aircraft$) +10000 * fast-burn($aircraft$) / refuel-rate($aircraft$))) ∨
　　// If zoom is impossible across the given distance.
　　([t] distance($city1$, $city2$) * fast-burn($aircraft$) > capacity($aircraft$)) ∨
　　// If zoom has to refuel immediately but fly does not.
　　([t] fuel($aircraft$) >= distance($city1$, $city2$) * slow-burn($aircraft$) ∧
　　 fuel($aircraft$) < distance($city1$, $city2$) * fast-burn($aircraft$))

### 5.2.4 ZenoTravel: Discussion

Finding control rules that yield good (but usually suboptimal) plans is not too difficult in the ZenoTravel domain. There are no risks involved in flying a plane to pick up passengers since all the passengers will always fit in the plane and refueling is possible in any city. In other words, it is not really possible to get stuck while looking for a solution. Also, since the graph of cities is fully connected, no route planning is necessary.

A fourth version of ZenoTravel, called Numeric, was available in the contest but due to lack of time we decided not to compete in this domain.

Among other things, the numeric version contains an additional constraint on the number of passengers that an aircraft can carry. At a first glance, this constraint may seem to introduce new problems. However, it is only enforced in the zoom operator, and since the numeric domain does not make use of durational operators, it suffers from the same problem as the STRIPS domain: The zoom operator consumes more fuel and limits the number of passengers, but does not deliver any advantages because it is no faster than flying.

The real difficulty in the Numeric version comes from the use of problem-specific metrics that measure the quality of a solution. For example, for one problem the planner may be required to minimize total-time + 3 * total-fuel-used, while for another problem it may be required to minimize total-time only. Until now, we have usually been satisfied with finding plans of good but not optimal quality, and this has been done by tuning control rules, for example by introducing the use-fly-instead-of-zoom function to determine whether fly or zoom should be used, as discussed above. This tuning is naturally done on the domain level rather than the problem level. An optimizing version of TALplanner is under development.

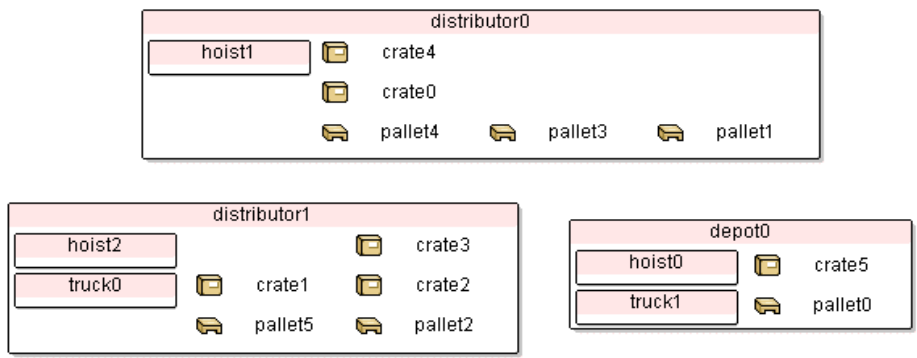

Figure 5: A Depots problem instance (STRIPS problem 7)

### 5.3 The Depots Domain

The Depots domain (illustrated in Figure 5) contains locations, trucks, hoists, movable crates, and pallets whose locations are fixed. Trucks move crates between any two locations and can carry any number of crates at the same time. Hoists are distributed among the locations and load crates into trucks or stack crates on surfaces (pallets or other crates). The goal is always to bring the crates into a certain configuration of stacks, where each stack is placed on a specific pallet.

**STRIPS.** The Depots domain is a combination of two other well-known planning domains, the logistics domain and the blocks world. Therefore it seems natural to start by taking a look at existing control rules for those two domains, and to see whether those rules can be combined easily or whether more complex rules are required due to interactions between moving and stacking blocks.

We begin with the blocks world part of the problem. The unbounded blocks world was used as a benchmark domain in IPC-2000, and there TALplanner used a modified version of the rules in Bacchus and Kabanza (2000) which ensure that the planner only adds blocks to "good towers", stacks that are already in their final position and will not have to be dismantled later in order to remove a block at a lower level. Can these rules be reused in the Depots domain? One prerequisite is the availability of temporary storage for all crates, since in the worst case every single stack of crates must be torn down completely before it is possible to start stacking crates on top of each other. Fortunately, although there is only a limited number of pallets, trucks can (somewhat counter-intuitively) contain any number of crates, and the planner can use them as storage. Only minor changes were required in order to handle the two separate types of surfaces: Pallets and crates.

Continuing with the logistics part, one simple rule can be reused from the standard logistics domain: Only unload a crate at its goal location. Its dual rule, "only load a crate if it needs to be moved", is not required. The blocks world rules ensure that a hoist does not lift a block unless it needs to be moved, and therefore it is already impossible to load such blocks into a truck.

It remains to ensure that vehicles only drive to those locations where they can be of use. In the standard logistics domain, a truck can drive to another location if there is a package that needs to be picked up or delivered there, but due to the use of stacks of crates in the depots domain, the rule must be modified: A vehicle may drive to a location if (1) there is a crate there that must be moved to another location, (2) there is a crate there that must be stacked differently, or (3) there is a crate in the truck that needs to be at the location, its destination is ready, and there is no other crate that should also be at the same location that the truck has not yet picked up.

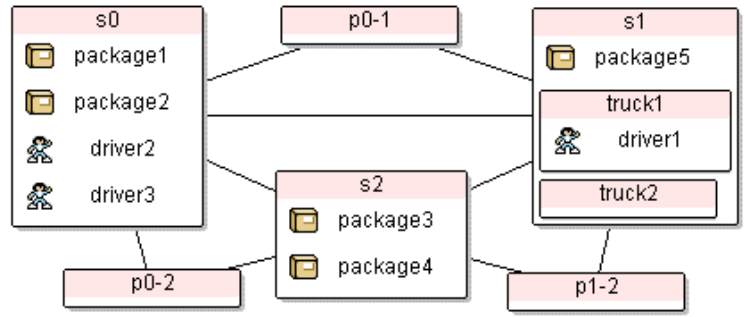

Figure 6: A DriverLog problem instance (STRIPS problem 5)

**SimpleTime.** In the SimpleTime version, lifting and dropping crates still takes one unit of time, loading takes three units, unloading four, and driving ten. A few changes were made to ensure mutual exclusion. For example, hoists can only lift one crate at a time. Also, a driving-to fluent was introduced to keep track of where trucks are headed, similar to flying-to in ZenoTravel.

**Timed.** In the Timed domain, the time required for loading and unloading a crate depends on how powerful the hoist is and on the weight of the crate. The time required for driving between two locations depends on the speed of the truck and the distance between the locations. Again, only minor changes were required to handle the domains, although higher quality plans could certainly have been produced by taking timing into account when determining which hoists and trucks to use.

### 5.4 The DriverLog Domain

DriverLog (illustrated in Figure 6) is yet another logistics domain, this time introducing the concept of truck drivers and road maps. A number of packages are transported between locations by trucks. There are two sets of routes connecting the locations: Links, where trucks travel, and paths, which drivers can walk along when not driving a truck. A truck can only have one driver at a time but can load as many packages as is needed.

**STRIPS.** Several control rules used in previous logistics domains were useful for DriverLog with minor modifications. For example, packages should only be loaded into trucks if they need to be moved, and should not be unloaded until they have reached their final destination.

On the other hand, a number of changes were necessary due to the use of road maps. Most importantly, vehicles were previously only allowed to drive to locations that were immediately useful because there were packages to be picked up or delivered. In the DriverLog domain there may only be direct roads between *some* locations (specified by a predicate link(*from*, *to*)), and a truck may have to move through several intermediate locations in order to reach its destination. Consequently the control rules must be relaxed to allow trucks to visit locations that are not useful in themselves. Nevertheless, some degree of goal-directedness is still required. One possible method is to identify for each vehicle the set of locations where the vehicle might be useful, and to require that it chooses one such location and then takes the shortest path to its chosen destination. This method was used in the competition with the help of the built-in shortest path algorithm discussed in Section 4.8 and a control rule stating that each step (each invocation of drive or walk) must decrease the distance to the current destination. The following definitions will be explained below:

#define [t] **reasonable-truck-location**(*truck*, *location*):
   // Omitted due to space constraints

#distfeature **driving-distance-between**(*from*, *to*) :domain integer :link link

#mindistfeature **driving-distance-to-location-satisfying-formula**
   :distfeature driving-distance-between :domain integer

#define [t] **driving-distance-to-reasonable-destination**(*truck*, *location*):
   driving-distance-to-location-satisfying-formula(*location*, *to*,
      [t] reasonable-truck-location(*truck*, *to*))

A boolean fluent reasonable-truck-location(*truck*, *loc*) is defined in terms of a logic formula, which specifies whether the given location is a reasonable destination for a given truck at the timepoint when it is evaluated. The driving-distance-between function accesses the shortest path algorithm to find the length of the shortest path between *from* and *to*, given that the road links are specified by the link predicate. The driving-distance-to-location-satisfying-formula function accesses another version of the shortest path algorithm and is used in driving-distance-to-reasonable-destination in order to find the shortest distance from *location* to any location *to* that satisfies reasonable-truck-location. Since all links have the same cost, it is then sufficient to require that whenever a truck moves, its driving-distance-to-reasonable-destination decreases.

Further changes were required due to the use of drivers. There may not be drivers for all trucks, so packages should not be loaded into a truck until the planner knows the truck will have a driver. Drivers should not disembark if there are still packages in the truck, or if there is a goal that the truck must be somewhere else. Drivers may have to walk along paths in order to reach a truck, so just like trucks, drivers must select one useful destination and then take the shortest path to their chosen destinations.

Additional control rules ensure that multiple trucks do not choose the same destination unnecessarily, and that multiple drivers do not choose to walk to the same location.

**SimpleTime.** In the SimpleTime version, loading and unloading objects takes two units of time, driving takes ten units, and walking takes twenty units. The operators are changed accordingly, and a going-to fluent is introduced to keep track of drivers and trucks that are moving towards a new location but have not yet arrived. A few minor adjustments must be made to the control rules.

**Timed.** In the Timed version, the time required to walk or drive between two locations is determined by a pair of functions specified in each problem instance. Since individual road segments can have different lengths, the method we used to ensure drivers and trucks used the shortest path to their current destination is no longer sufficient, and must be modified slightly. Other than this, there are no major changes for the Timed version.

### 5.5 The Rovers Domain

The Rovers domain simulates a simplified planetary exploration expedition. A lander vessel carries a number of rovers to the planet surface and provides a communication link back to Earth. Each rover has a subset of the general capabilities, retrieving soil samples, retrieving rock samples and capturing images using cameras that support different imaging modes. The cameras are mounted on the rovers, as are storage compartments, one for each rover, which can hold one soil sample or one rock sample. Data from a sample must be sent to the lander by a communication link. All missions revolve around navigating waypoints on the planets surface to collect samples and take images of specified objectives that are only visible from certain waypoints. The terrain may prevent rovers from going directly between two waypoints and different rovers handle different terrain so a list of routes each rover can use is provided.

**STRIPS.** Following a control scheme similar to the one used in DriverLog, we limit the movements of rovers to locations where they can perform some useful action like collecting a rock sample or capturing an image. The problem of finding a path from one waypoint to another is also solved in the same way as in DriverLog, except that each rover has its own set of routes between waypoints.

**SimpleTime.** The changes in the SimpleTime version are trivial: Operator durations are changed, a few mutual exclusion relations need to be enforced, and a new fluent calibrating(camera) keeps track of whether a certain camera is being calibrated.

**Timed.** The Timed version introduces the concept of energy, where each rover has a limited amount of energy and each action it does consumes some of the energy. This is similar to the use of fuel in the ZenoTravel domain, but there is also a major difference: The rovers have been equipped with solar panels that recharge the rover, but only some of the waypoints that a rover can go to are directly exposed to the sun, which is a requirement for the solar panels to work. The airplanes in the ZenoTravel domain can refuel anywhere, and so fuel usage is only relevant in terms of minimization of resource usage, whereas a rover that uses its energy unwisely can get stuck in the shade, unable to do anything or go anywhere. To prevent this we can either let the planner backtrack and search for a better plan, or we can introduce stricter rules that keep energy levels in mind when deciding what a rover is allowed to do. The latter approach is taken below.

The critical point is when a rover does not have enough energy to reach a waypoint in the sun and recharge. Using the shortest path algorithm it is possible for a control rule to determine the distance to the closest waypoint that is exposed to the sun. In addition to all waypoints that were previously allowed, it is also reasonable for a rover to go to a waypoint that is exposed to the sun if the rover does not have enough energy to perform an action and then go recharge, or if there do not exist any other waypoints that are both affordable and reasonable to visit.

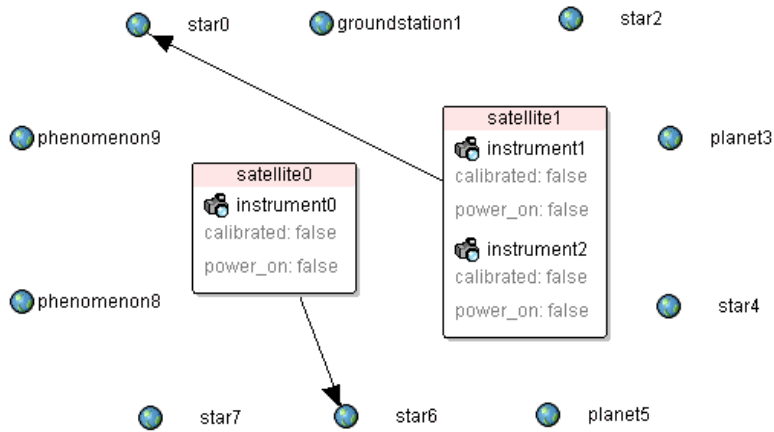

Figure 7: A Satellite problem instance (STRIPS problem 4)

### 5.6 The Satellite Domain

In the Satellite domain a number of satellites orbit the Earth, each equipped with a set of scientific imaging instruments. The satellites turn in space, targeting stars, planets and interesting phenomena to capture images of them using different instrument operation modes. These modes can include regular or infrared imaging and spectrographic or thermographic readings but are different for each problem. The planner's task is to schedule a series of observations so that the satellites are used efficiently. Figure 7 shows a small example problem instance, with arrows showing the directions in which the satellites are pointing.

Directions are not represented as explicit coordinates. Instead, satellites can turn to a new direction by giving the turn_to operator an argument specifying the star, planet or phenomenon that the satellite should point to. Instruments first need to be activated using switch_on, then calibrated at a calibration target with the calibrate operator before they can capture images using take_image. Each satellite has only enough power to operate one instrument at a time, so switching active instruments is always initiated by the switch_off operator to deactivate the first instrument.

#### 5.6.1 Satellite: STRIPS

Since the task consists of collecting a number of images, we begin by restricting the use of take_image to images that are mentioned in the goal.

#control :name "**only-take-pictures-of-goals**"
   $[t]$ ¬have_image(*direction*, *mode*) ∧ $[t+1]$ have_image(*direction*, *mode*) →
   goal(have_image(*direction*, *mode*))

The next step is to restrict the directions in which satellites turn to those that may actually help in collecting the images. The task is split into a control rule, only-point-in-goal-directions, and a definition of goal directions. A satellite is allowed to turn towards a direction to take a picture, to calibrate an instrument or if a goal specifies that the satellite should point in the direction and there is no more work left to do.

#define [t] **goal_direction**(*satellite*, *direction*):
   [t] take_image_possible(*satellite*, *direction*) ∨
   ∃*instrument* [
      [t] power_on(*instrument*) ∧ ¬calibrated(*instrument*) ∧
      [t] calibration_target(*instrument*, *direction*) ∧ on_board(*instrument*, *satellite*) ] ∨
   goal(pointing(*satellite*, *direction*)) ∧ [t] all_images_collected

The take_image_possible function checks not only if an image is to be collected but also that it has not already been taken and that the satellite has the necessary instrumentation ready. If the active instrument is not calibrated, the satellite may first have to turn towards another direction and calibrate it.

#define [t] **take_image_possible**(*satellite*, *direction*):
   ∃*mode* [ goal (have_image(*direction*, *mode*)) ∧
         [t] ¬have_image(*direction*, *mode*) ∧
         ∃*instrument* [
            [t] power_on(*instrument*) ∧ calibrated(*instrument*) ∧
            [t] on_board(*instrument*, *satellite*) ∧ supports(*instrument*, *mode*) ]]

The switch_on and switch_off operators are still not regulated by control rules and the planner quickly takes up the habit of repeatedly flipping the power to different instruments on and off. Once an instrument has been powered on and calibrated, using it as much as possible before switching to another instrument seems reasonable. A usefulness function, putting a value on the usefulness of a particular instrument, helps decide which instrument to power on first.

#define [t] **usefulness**(*instrument*):
    value(t, $sum(<*mode*>, [t] supports(*instrument*, *mode*) ∧ mode_needed_for_goal(*mode*), 1))

#define [t] **mode_needed_for_goal**(*mode*):
     ∃*direction* [ goal(have_image(*direction*, *mode*)) ∧ [t] ¬have_image(*direction*, *mode*) ]

Add one to the usefulness score of an instrument for each imaging mode that it supports and that is needed in some goal. This score is then used in a control rule that chooses a satellite's most useful instrument, if it has any.

#control :name "**use-the-most-useful-instrument**"
   [t] ¬power_on(*instrument*) ∧ [t+1] power_on(*instrument*) →
   [t] usefulness(*instrument*) > 0 ∧
   ¬∃*satellite*, *instrument2* [
      [t] usefulness(*instrument2*) > usefulness(*instrument*) ∧
      [t] on_board(*instrument*, *satellite*) ∧ on_board(*instrument2*, *satellite*) ]

Switching off an instrument is only allowed if the instrument is no longer required.

#control :name "**don't-switch-instrument-off-if-you-don't-have-to**"
   [t] power_on(*instrument*) ∧ [t+1] ¬power_on(*instrument*)) →
   [t] ¬∃*mode* [ supports(*instrument*, *mode*) ∧ mode_needed_for_goal(*mode*) ]

We have run out of more or less obvious improvements, but analyzing the planner output reveals one remaining inefficiency: The satellites often simultaneously decide to turn to the same direction because a picture needs to be taken in that direction, despite the fact that only one satellite needs to take the picture. This is similar to the situation in the ZenoTravel domain where a number of aircraft may concurrently choose to pick up the same passenger, but there are some differences due to the fact that the *only* reason for a satellite to point in a certain direction is in order to calibrate itself or take an image, which makes the task somewhat easier.

Therefore this problem can be solved in a different way, using a resource for mutual exclusion. This resource, called point_towards(*direction*) and having a capacity of 1, can be borrowed temporarily by turn_to for the duration of the turn. If one satellite turns towards a specific direction $d$, no other satellite can turn towards $d$ without causing a resource conflict.

This still leaves one problem: When the first satellite has finished turning, it no longer owns the point_towards($d$) resource and therefore another satellite can immediately start turning towards $d$. It is no longer possible for more than one satellite to turn towards the same direction at once, but while the first satellite is taking pictures, other satellites can turn to that direction one by one, until finally all the desired pictures have been taken in that direction and goal_direction sees that there is no longer any valid reason to point towards $d$. This can be solved either by changing the definition of goal_direction or by letting take_image borrow the same resource.

Clearly, this type of "swarming" problem occurs quite often in concurrent domains and a more principled solution should be investigated in the future.

### 5.6.2 Satellite: SimpleTime

The SimpleTime version changes the duration of some operators. Turning takes five time units, switching an instrument on takes two units, calibrating it takes five units and taking a picture takes seven units. A couple of helper fluents, turning_towards, calibrating, have_image_generalized (an image exists or is being taken) and power_on_generalized (power is on or a switch_on action is being executed) keep track of actions that have begun but not completed. The affected control rules are updated accordingly.

### 5.6.3 Satellite: Timed

The Timed version of the Satellite domain includes two new functions. The calibration_time specifies the time required to calibrate, while the slew_time function represents the time required for a satellite to turn between two directions. Neither of these changes prompts any significant changes to the SimpleTime control.

### 5.6.4 Satellite: Discussion

The Satellite domain does not provide a real challenge as long as the planner is only trying to find a correct plan. Finding a short plan is harder, especially in the Timed version, and would require additional analysis to determine in which order images should be collected and which satellites should be used for each image. Doing this using control rules seemed a bit like overkill, especially since we had not yet created control rules for the complex UMTranslog-2 domain. For this reason, we decided to be satisfied with what we had done so far, and were surprised when the plans we generated turned out to be of considerably lower quality than those produced by some other planners.

After the contest, we were informed of the reason, or at least the main reason: The automatic problem generator that created the problem instances randomized the slew times between every pair of directions and did not check for geometrical consistency that would be present in a real world situation. We had subconsciously assumed that the problem instances satisfied the triangle inequality, but this was not the case, and the other planning teams had discovered this. For example, in handcoded problem 14, turning a satellite directly between phenomenon86 and groundstation4 takes 82.860 units of time, while turning it through two carefully selected intermediate directions requires 1.183 units of time.

Initial testing shows that taking this into consideration and once again using the built-in shortest path algorithm yields significantly shorter plans when plan length is measured by the time point at which the goals have been satisfied.

Another potential improvement would be to change the last clause in goal_direction to allow satellites to turn towards a direction specified in the goals as soon as one has started taking the last picture, rather than waiting until one has finished taking the last picture.

### 5.7 The UMTranslog-2 Domain

The UMTranslog-2 domain is another logistics domain, but with 14 types, 38 predicates, 24 functions and 38 operators, its size and complexity is incomparable to the previously encountered logistics domains in the contest.

Since the formal domain definition was the only information provided about the domain and there was no high-level description, we had to work out all the information about the domain from the PDDL definition. This was not a major problem for the previous domains, since they were generally quite simple and easy to understand, but it did give us some problems in UMTranslog-2. A significant amount of time was spent trying to determine exactly how packages were allowed to move and how they can be loaded into and unloaded from various kinds of vehicles. In retrospect, it would probably have been better to do as some other teams did: Skip the UMTranslog-2 domain completely and spend that time on the Numeric and Complex versions of the other domains.

**The domain.** Trucks, trains or aircraft transport packages between locations but they must follow strict movement patterns. A few locations are transportation hubs, some are transportation centers while the rest are ordinary locations. A package is only allowed to move up and down through this hierarchy once and only move between two locations in the same layer once. The longest possible route for a package is thus from an ordinary location to a transportation center to a hub to another hub to a transportation center and finally to another ordinary location.

The domain groups locations into cities, which are then grouped in regions. Trucks travel between any two locations in the same city or by an existing road route between two cities. Trains and planes always use predefined routes between transportation centers and hubs. A great number of restrictions further complicate movements. Packages must be compatible with the vehicle they are loaded into, the vehicle must have enough free space, not be loaded too heavily and not be wider, longer or higher than the route and destination location accepts. Finally, the locations, vehicles and routes must all be available for use.

**Control rules.** As in previous domains, we specify what a reasonable location is and limit vehicle movements to destinations that are reasonable. A truck might want to pick up or deliver a package at the location or, if the truck cannot reach the goal location of the package, unload the package at a transportation center to be picked up by another vehicle. Our control rules do not allow trucks to pick up several packages. This makes finding optimal solutions impossible in the general case but simplifies the search for acceptable solutions a great deal. There is an imminent risk that any other packages the truck is carrying will end up at the wrong location if it is allowed to travel about, picking up more packages along the way. Since all packages must move according to the specified pattern of transportation centers and hubs, moving a package that has once arrived at a location that is not a transportation center is not allowed and the package will be stuck there. Restricting trucks to picking up one package at a time avoids this problem.

There is also a large group of loading and unloading rules controlling, among other things, the opening or closing of valves and doors and loading or unloading of packages. Finally, packages are only loaded into vehicles that are actually able to take them to a useful location.

Creating control rules and meeting the contest deadline left no time to get the domain working with concurrent planning. Instead, we had to make do with sequential planning.

Given more time, the set of control rules could definitely be improved. If planning speed is less of an issue, more search can be allowed and higher quality plans generated. More and better problem instances would be needed as guidelines when developing better control rules since the contest problems did not make full use of the intended transportation scheme with transportation centers and hubs.

## 6. Improvements After the Competition

Though the planning competition ended during the AIPS-2002 conference in April, 2002, our work on TALplanner naturally did not cease there. There are still many improvements that can be made, and a couple of them that are related to the development of new domains and control rules have been implemented during the summer of 2002.

### 6.1 Domain Visualization

As was discussed in the description of the ZenoTravel domain, the process of creating control rules for a planning domain often involves incremental improvements. TALplanner is run on a number of problem instances using one set of control rules, or possibly without any control rules at all, and the beginning of the resulting search tree is analyzed in order to determine where bad choices were made and how they can be avoided using new or improved control rules. This is repeated until the planner consistently finds plans of good quality.

During this process, one must study not only the output of the planner but also the structure of the particular problem instance being solved. For example, in a DriverLog problem it may be necessary to draw the road network being used in each problem instance using pen and paper, and then study the paths taken by trucks, people, and packages through the road network, in order to discover whether improvements would be possible. But often a particular inefficiency only appears in one or a few out of a large set of problem instances, and tracing the execution of each plan by hand is obviously a tedious and time consuming task that ought to be automated as far as possible.

This led to the development of TPVis, a generic graphical visualization framework for TALplanner. The TPVis framework was used to generate the domain images in this article, and provides an animated display consisting of a set of nodes, where each node can be a container or an atomic object. Containers may represent vehicles (which can contain packages), locations (since there can be vehicles, packages or other objects at a location) or other similar concepts, while atomic nodes may be used for packages, instruments on a satellite, or any other type of object which should be displayed. Edges between nodes can indicate any form of relationship between objects, the most obvious interpretation being that two location nodes are connected by some transportation route. A built-in layout engine can generate a layout automatically, or you can manually adjust the visual coordinates of each node.

The visualization framework is then used by concrete plugins adapted to specific planning domains. The DriverLog plugin, for example, displays locations as container nodes, linked by paths where drivers can walk and links where trucks can drive. Trucks are also containers, contained within a specific location, as shown in Figure 6 on page 366.

As a plan is being generated, TPVis animates the actual movements of objects between locations. This creates a better instinctive feel for the domain, and the two-dimensional graph display gives an overview that is difficult to provide using only text output. In addition to animating a graph, TPVis simultaneously lists the partial plan leading up to the current state and the problem goals that the planner tries to satisfy. TPVis also provides a limited form of interactive planning since it, at any point in the planning process, allows the user to force the planner to backtrack and explore a different search branch.

The development of TPVis was not initiated until after the planning contest. If this graphical visualization had been available during the work on the contest domains, it would have saved a lot of time, and possibly a tree or two.

### 6.2 Automatic translation from PDDL to TALplanner

Although it was obvious that there should be an automatic translator from PDDL to TALplanner's input format, there were always more urgent features to be implemented, and we instead decided to translate the IPC-2002 domains by hand. In retrospect this was a mistake. The risk of making an error somewhere in the translation becomes imminent when dealing with complex domains such as UMTranslog-2, with 38 operators, some of which had highly complex preconditions. Also, translating long formulas by hand is quite time consuming. A semi-automatic translator was therefore implemented to decrease the amount of work involved in the translation process and reduce the risk of introducing errors in the definition.

## 7. Discussion and Conclusions

The third International Planning Competition was a major step forward in terms of the expressibility required to represent the benchmark domains, and it provided a number of interesting challenges for any planner that wanted to participate in the competition. In this article we have described how these challenges affected TALplanner and shown a number of extensions that were made in order to meet the challenges. The article also includes a number of domain-dependent control rules for the competition domains, but rather than presenting an exhaustive list of pre-packaged control rules, we have attempted to place more emphasis on explaining the incremental analysis process that eventually leads to the final formulas, going into particular detail for the ZenoTravel domain.

As could be seen in the examples shown in this paper, control rules are often simple, natural common-sense rules, and not very difficult to generate given some basic knowledge about the planning domain. Some rules are more complex, but still not difficult to understand or verify once someone has spent the effort to generate them. And then, unfortunately, there are a few rules that are quite unintuitive, rules that are too complex to be easily understood, and rules that occasionally forbid optimal plans.

To some extent, such rules might be avoided by gaining more experience in good practices for writing control rules, or by extending the expressivity of the language in which control rules are written so that complex conditions can be expressed more succinctly or in a more natural manner, or simply by spending a little bit more time on the control rules than was available during the planning competition when much of our time was spent teaching or working on the planner implementation. However, another important cause for the complexity of certain rules is probably that we are attempting to express all search control knowledge in the same way: As control rules that prune the search tree to such a great extent that even a simple depth-first search algorithm is sufficient for efficiently finding good plans in the remainder of the tree.

Not all search control knowledge can easily be expressed in this manner, but this certainly does not mean that control rules should be abandoned altogether. Instead, what we learn from this experience is that control rules might not be the one and only multi-purpose planning tool that will efficiently and easily solve all our planning problems. Instead, just like one would expect, they are one very useful tool that deserves a place in our toolbox but should be combined with other approaches to planning. Just to mention one rather obvious example, it would be possible to devise a heuristic forward-chaining planner whose search tree would be pre-pruned using control rule techniques from TALplanner. Control rules could be written to exclude plans where the heuristic gives a suboptimal result, potentially providing plans that are closer to optimal, and even for domains where the heuristic search function provides good plans it may often be more efficient to state a number of constraints as explicit control rules.

Such extensions to TALplanner have been considered at least since some time before the second planning competition in 2000, and it has long been clear to us that this approach should eventually be examined and explored. Before we could start working on this, though, the strengths and weaknesses of control rules had to be explored in more depth. Up to now, our work has therefore focused mostly on investigating how far it is possible to take TALplanner in its current shape, with explicit control rules being the only means for controlling the search process. This work has proved rather fruitful in itself, and TALplanner did well in IPC-2000 as well as in IPC-2002. The planner is now becoming reasonably mature, and after a few more improvements have been made and the planner has been released for general use, it might be time to take a step back and consider its relation to other approaches in more depth than has been done previously in order to investigate the possible advantages of hybrid approaches.

Of course, this does not mean that there is nothing more to be done within the "pure" TALplanner framework. On the contrary, there are many additional topics to be pursued, including investigating the application of TALplanner to plan optimization problems (where the very simplest approaches might involve applying standard optimal graph search algorithms to the pruned search tree generated by TALplanner) and extending the planner to handle incomplete knowledge and non-deterministic operators. Which of these many topics will be the next focus of our research has not yet been determined.

## Acknowledgements

This research is supported in part by the WITAS Project under the Wallenberg Foundation.